\documentclass[lettersize,journal]{IEEEtran}
\usepackage{amsmath,amsfonts}
\usepackage{algorithmic}
\usepackage{algorithm}
\usepackage{array}
\usepackage[caption=false,font=normalsize,labelfont=sf,textfont=sf]{subfig}
\usepackage{textcomp}
\usepackage{stfloats}
\usepackage{url}
\usepackage{verbatim}
\usepackage{setspace}
\usepackage{graphicx}
\usepackage{cite}
\usepackage{color}
\usepackage[colorlinks,linkcolor=blue]{hyperref}

\begin{document}

\title{MST: Adaptive Multi-Scale Tokens Guided Interactive Segmentation}

\author{
\IEEEauthorblockN{Long Xu\textsuperscript{1}},
\IEEEauthorblockN{Shanghong Li\textsuperscript{1}},
\IEEEauthorblockN{Yongquan Chen\textsuperscript{1}\IEEEauthorrefmark{1}},
\IEEEauthorblockN{Jun Luo\textsuperscript{2}},
\IEEEauthorblockN{Shiwu Lai\textsuperscript{3}}
\thanks{
\noindent\textsuperscript{1} Shenzhen Institute of Artificial Intelligence and Robotics for Society, The Chinese University of Hong Kong, Shenzhen, China.

\textsuperscript{2} School of Information Science and Engineering, Northeastern University, Shenyang, China.

\textsuperscript{3} Maxvision Technology Corp., Shenzhen, China.

\IEEEauthorrefmark{1} Corresponding author: Yongquan Chen (e-mail: yqchen@cuhk.edu.cn).
}

\thanks{Manuscript received April 19, 2021; revised August 16, 2021.}}

\markboth{Journal of \LaTeX\ Class Files,~Vol.~14, No.~8, August~2021}%
{Shell \MakeLowercase{\textit{et al.}}: A Sample Article Using IEEEtran.cls for IEEE Journals}


\maketitle

{\setlength{\parskip}{1.0\baselineskip
\textcolor{red}{This work has been submitted to the IEEE for possible publication. Copyright may be transferred without notice, after which this version may no longer be accessible.}}}

\begin{abstract}
Interactive segmentation has gained significant attention for its application in human-computer interaction and data annotation. 
To address the target scale variation issue in interactive segmentation, a novel multi-scale token adaptation algorithm is proposed.
By performing top-k operations across multi-scale tokens, the computational complexity is greatly simplified while ensuring performance.
To enhance the robustness of multi-scale token selection, we also propose a token learning algorithm based on contrastive loss. This algorithm can effectively improve the performance of multi-scale token adaptation.
Extensive benchmarking shows that the algorithm achieves state-of-the-art (SOTA) performance, compared to current methods.
An interactive demo and all reproducible codes will be released at \url{https://github.com/hahamyt/mst}.

\end{abstract}


\section{Introduction}
\IEEEPARstart{T}{he} interactive image segmentation algorithm, which can generate semantic masks for images with just a few clicks, has garnered widespread attention in recent years \cite{chen2022focalclick,liu2022simpleclick,9897365}.

However, existing algorithms face performance bottlenecks due to the target scale variation issue.
For example, in remote sensing segmentation task \cite{loveda}, targets like green land and water systems are typically much larger than buildings and roads, making it difficult to handle using single-scale features.

\begin{figure}[!t]
    \centering
    \includegraphics[scale=0.76]{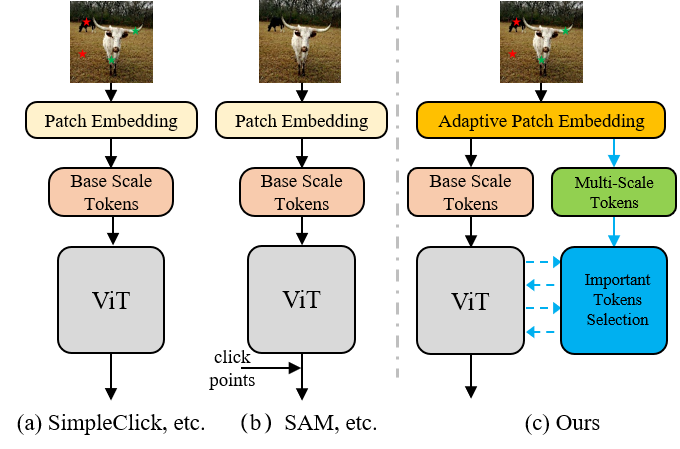}
    \caption{The differences between the proposed algorithm and the existing methods, including the processing of click points and the use of multi-scale features}
    \label{fig:motivation}
\end{figure}

To address this issue, Liu et al. \cite{liu2022simpleclick} used the SimpleFPN proposed by He et al. \cite{li2022exploring} to capture the scale variations of the ViT \cite{dosovitskiy2020image} features, which improved the segmentation accuracy.
This structure was also adopted in many subsequent works \cite{sun2023cfr,lin2023adaptiveclick}.
However, this strategy simply interpolated the single-scale ViT \cite{dosovitskiy2020image} features into different scales.
But it was difficult to inherently alleviate the limitations of single-scale features.

To obtain better multi-scale features, Fan et al. \cite{fan2021multiscale} proposed a multi-scale vision transformer model using pooling attention across space-time resolution and channel dimension.
This algorithm used different blocks to characterize multi-scale features, which is also adopted by Segformer \cite{xie2021segformer}.
Chen et al. \cite{chen2022focalclick} introduced Segformer's four-stage blocks as image encoders for interactive segmentation, achieving better accuracy by utilizing equivalent multi-scale features.
However, this multi-scale feature is still a sampling of the features from the previous stage, making it difficult to directly obtain multi-scale features from the input.

In addition to interactive segmentation using multi-scale transformers, Sofiiuk et al. \cite{9897365} and Chen et al. \cite{chen2021conditional} adopted CNN-based HRNet \cite{wang2020deep} and ResNet \cite{he2016deep} to capture robust multi-scale convolution features, which helped the algorithms perform better in target scale variation scenarios.
However, these algorithms usually use the final output of the backbone, although they can capture the semantic information of the target well, it is difficult to obtain robust multi-scale information, which limits their accuracy and robustness for target scale variations.

\begin{figure}[!t]
    \centering
    \includegraphics[scale=0.71]{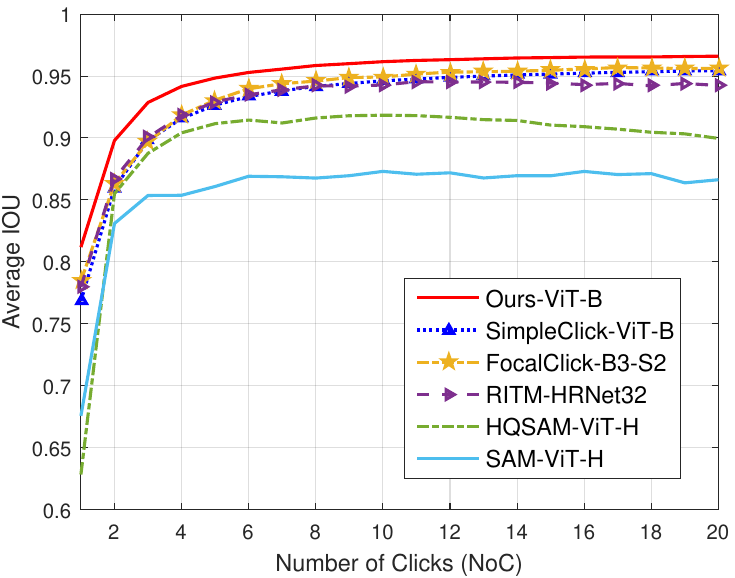}
    \caption{The average IOU varies with clicks (based on the average results of 6 benchmarks), indicating the proposed method can utilize fewer clicks to obtain better precision}
    \label{fig:miouclick}
\end{figure}

Although these works can alleviate the target scale variation issue to some extent, they have ignored the effective use of multi-scale features in the input stage.

To address this issue, we proposed a multi-scale token fusion strategy, which adaptively selects important multi-scale tokens as a query set to guide the base ViT tokens to pay more attention to important local regions.
To identify the important tokens, the system selects those multi-scale input tokens that are most similar to the base tokens which have been positively clicked.

Moreover, to improve the robustness of important token selection, the contrastive loss \cite{he2020momentum} is introduced to enhance the discriminability of tokens.
Unlike existing algorithms, the proposed algorithm directly obtains multi-scale tokens at the input stage, which improves the performance in multi-scale target segmentation.

Furthermore, the proposed important token selection algorithm and contrastive loss can fully utilize the user interaction information, which is vital in human-computer interaction.

In summary, the proposed algorithm contributes the following innovations to interactive segmentation:
\begin{itemize}
    \item A similarity-based multi-scale token interaction algorithm is introduced, improving the performance of fine-grained segmentation.
    \item A token learning algorithm based on contrastive loss is proposed, which enhances the discrimination between positively clicked tokens and background tokens.
\end{itemize}

\section{Related Work}

The proposed algorithm primarily focuses on utilizing input multi-scale tokens in ViT \cite{dosovitskiy2020image}.
As such, related work will be mainly discussed regarding the multi-scale vision transformer and important token selection.

\subsection{Multi-scale Vision Transformer}
Chen et al. \cite{chen2021crossvit} pointed out that smaller patch sizes can bring better performance in vision transformers, but smaller patches also lead to longer token lengths and higher memory requirements. 
To mitigate the memory and computation issues, they used small and large patch sizes to generate tokens and fused these two sizes of tokens via cross-attention.
Compared to computing only small patches, incorporating large patches and cross-attention significantly reduces complexity while maintaining the performance advantages of small patch sizes.
However, this strategy requires all tokens to participate in the calculation, increasing computation and reducing speed.

Different from using input stage multi-scale tokens algorithms, Xie et al. \cite{xie2021segformer} proposed a scaled self-attention algorithm, which down-sampled the key and value into larger size tokens using a scale ratio.
This algorithm’s self-attention calculation is equivalent to the cross-attention between the base tokens and larger tokens.
However, this strategy can only obtain larger tokens, and cannot produce tokens smaller than base tokens.  

Similar to \cite{xie2021segformer}, Ren et al. \cite{ren2022shunted} also introduced a multi-scale self-attention strategy. 
Unlike \cite{xie2021segformer}, which only used a single scale ratio, Ren et al. adopted two different scale ratios to downsample the base tokens, incorporating more abundant scale information from the input.
However, this strategy still cannot capture smaller tokens of the input. 
Additionally, its complex structure introduces substantial computation overhead.

\subsection{Important Token Selection}
The computational complexity of the transformer is quadratically dependent on token length \cite{dosovitskiy2020image}.
In recent years, many transformer-based image processing methods have been proposed to simplify the task.

Dosovitskiy et al. \cite{dosovitskiy2020image} first used 16$\times$16 pixels image patch in Vision Transformer (ViT) to limit the token length.
Ranftl et al. \cite{ranftl2021vision} upscaled the low-resolution feature maps of ViT to high-resolution for dense prediction tasks.

These algorithms need to calculate all tokens in self-attention, which is computationally intensive in high-resolution input scenarios.

To efficiently select important tokens, 
Wang et al. \cite{wang2022efficient} proposed a score-based algorithm for selecting important tokens in video processing.
The algorithm adopted a differentiable top-k selection algorithm proposed by Cordonnier et al. \cite{cordonnier2021differentiable}.
Although this algorithm successfully obtained equivalent performance by using much fewer tokens, it requires a large amount of GPU memory when processing long sequences of tokens.
Additionally, it only selects base tokens, resulting in poor multi-scale performance.

Similar to \cite{ren2022shunted}, Tang et al. \cite{tang2022quadtree} proposed distinguishing important and unimportant tokens to skip unimportant regions and subdivide important areas.
This algorithm effectively directed fine-grained attention to important areas and significantly reduced computation.

However, an inherent limitation persists in reliance on a single input scale, with average pooling or cross-attention used to change the scale of tokens.
This restricts multi-scale modeling capability for capturing target changes.
Finally, as only selected tokens participate throughout the process, the loss of potentially important information remains a risk.

\begin{figure*}[!t]
    \centering
    \includegraphics[scale=0.73]{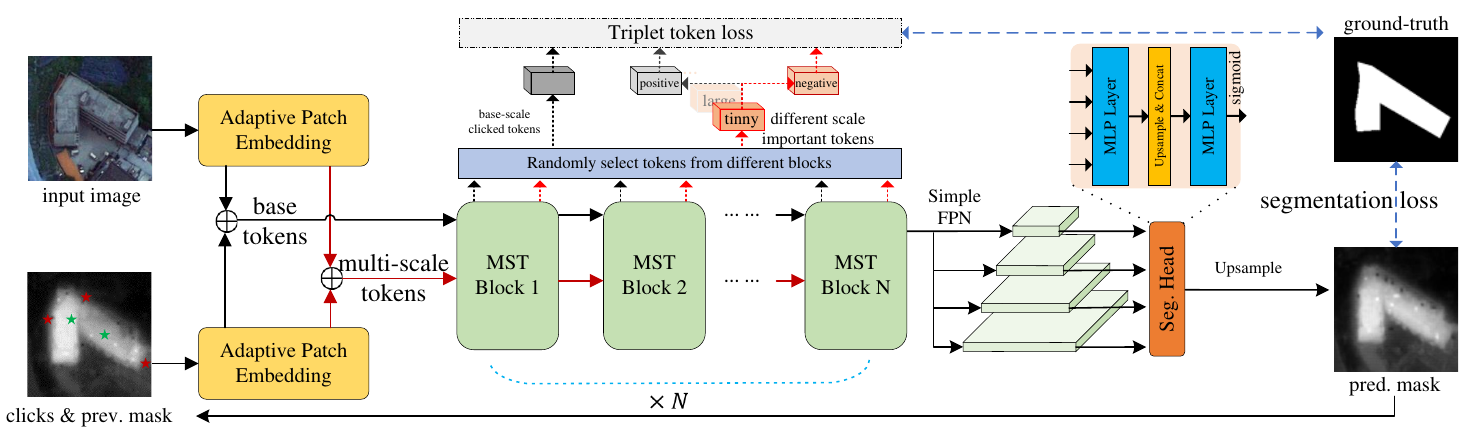}
    \caption{The overall framework of the proposed algorithm. The \textbf{adaptive patch embedding} module is adopted to extract multi-scale tokens adaptively, base tokens denotes the tokens with patch size 16$\times16$, multi-scale tokens denotes the tokens with patch size $8\times 8$ and $28\times 28$. \textbf{MST block} is the proposed multi-scale token fusion module, more details can be found in Fig. \ref{fig:multiscale}. For efficient training, we utilize a \textbf{random selection module}, which has no impact during inference. The \textbf{triplet token loss module}  represents our proposed contrastive loss-based token learning algorithm, more details can be found in Fig. \ref{fig:contrastloss}. A simple FPN is adopted in this paper, and segmentation is performed using a two-layer MLP module }
    \label{fig:framework}
\end{figure*}

\section{Method}
To improve multi-scale performance for interactive segmentation, we propose an adaptive multi-scale tokens fusion algorithm and a contrastive loss between target and background tokens.

The overall framework is illustrated in Fig. \ref{fig:framework}.
Adaptive Patch Embedding denotes the multi-scale token extraction module, which obtained different patch size tokens by convolution module with different interpolated weights.
In this paper, $8\times8$,$16\times16$, and $28\times28$ patch sizes are adopted.

In this paper, ViT is adopted as an image encoder, using $16\times16$ base tokens (\textbf{black} lines, Fig. \ref{fig:framework}).
The multi-scale tokens (\textbf{\textcolor{red}{red}} lines, Fig. \ref{fig:framework}) comprise additional $8\times8$ and $28\times28$ tokens.

The multi-scale tokens fusion (MST) module (Fig. \ref{fig:multiscale}) adopts the feature pyramid structure SimpleFPN  \cite{li2022exploring}, originally introduced for interactive segmentation in SimpleClick \cite{liu2022simpleclick}.

\subsection{Preliminaries}
The proposed interactive segmentation algorithm follows the standard framework used in most current approaches \cite{liu2022simpleclick, chen2022focalclick, 9897365}, except SAM \cite{kirillov2023segment}.
It includes input image $x_{in}\in\mathbb{R}^{B\times3\times W\times H}$, click maps $x_c\in\mathbb{R}^{B\times2\times W\times H}$, and previously generated mask $x_{m}=\text{sigmoid}(x_P)\in\mathbb{R}^{B\times 1\times W\times H}$,
where 2 in $x_c$ means the two channels representing positive and negative clicks. 
$x_P$ denotes the location distribution of the target, while $x_m$ denotes the final segmentation results.

As illustrated in Fig. \ref{fig:framework}, the predicted results $x_P\in\mathbb{R}^{\frac{W}{4}\times\frac{H}{4}\times N_{cls}}$ are obtained using a multilayer perceptron (MLP), with $N_{cls}=1$.

Standard ViT \cite{dosovitskiy2020image} divides $x$ into $L_b$ patches of $16\times 16$ pixels, generating corresponding tokens $f_{b}\in \mathbb{R}^{B\times L_b\times C}$ through Patch Embedding module.
Here, $B$ is the batch size, and $b$ denotes the base $16\times 16$ tokens. $C$ is the embedding dimension of tokens, while $f_b$ is named base tokens.

In addition to base tokens, this study also uses tokens with patch sizes $8\times 8$ and $28\times 28$, which are denoted as tinny tokens $f_t\in\mathbb{R}^{B\times L_t\times C}$ and large tokens $f_l\in\mathbb{R}^{B\times L_l\times C}$ respectively.
The proposed algorithm can be easily extended to include tokens with more patch sizes.

However, using smaller patch sizes generates more tokens, increasing computational complexity, as illustrated in formula (\ref{eq:selfattention1}).

\begin{equation}\label{eq:selfattention1}
    \begin{aligned}
        A = \text{softmax}\left(\frac{Q\cdot K^T}{\sqrt{d_k}}\right)
    \end{aligned}
\end{equation}
where $Q,K,V\in\mathbb{R}^{H\times L\times (C/H)}$, $A\in\mathbb{R}^{H\times L\times L}$, and $H$ denotes the number of heads.
To reduce the computation complexity of the attention mechanism, we propose a similarity-based differentiable top-k token selection algorithm, which only selects important tokens from multi-scale tokens to update base tokens.

\subsection{Multi-scale Tokens Selection and Interaction}

The base tokens $f_b$ are usually hard to handle with fine-grained target segmentation and are also not good for large targets.
To address this issue, this study introduces multi-scale tokens that interact with base tokens.
These tokens are adaptively selected based on the input target scales.
The main challenges of the scale adaption are computational cost and token selection strategy.

Inspired by \cite{he2022masked} and \cite{cordonnier2021differentiable}, we have found that tokens contain much redundancy information.
Using only important tokens for computation can achieve equivalent performance.

This work uses $8\times 8$ and $28\times 28$ patch-sized tokens to extract multi-scale information from input $x$.
The corresponding tokens are represented as $f_t$ and $f_l$, respectively.

A scale adaptive patch and position embedding algorithm \cite{flexvit} is implemented to share information among different-sized tokens.
To obtain important tokens, we propose a differentiable top-k selection algorithm based on the natural assumption that the tokens within the image area clicked by users hold the highest importance.

\begin{figure}[!t]
    \centering
    \includegraphics[scale=0.37]{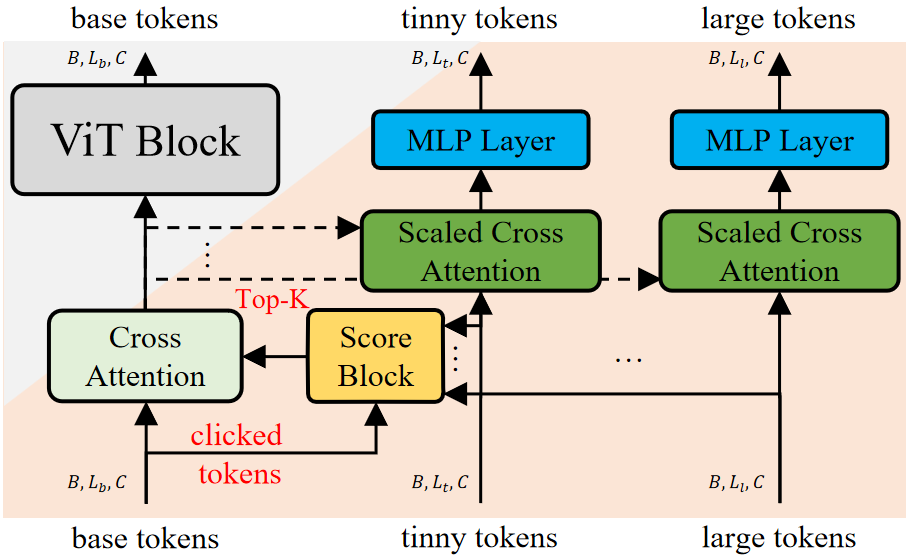}
    \caption{The proposed multi-scale tokens interactive module. The \textbf{ViT Block} denotes the original vision transformer block. The \textbf{Score Block} is the proposed algorithm for selecting important tokens.
    The \textbf{Cross Attention} is a feature fusion algorithm; further details can be found in \cite{fu2019dual}. The \textbf{Scaled Cross Attention} is an efficient cross-attention algorithm referenced in \cite{ren2022shunted}. The \textbf{Top-K} operation is a PyTorch function that selects the largest $K$ values in a vector. The term \textbf{clicked tokens} refers to the local region that has been clicked}
    \label{fig:multiscale}
\end{figure}

Based on this hypothesis, we use positively clicked tokens in $f_b$ as a reference kernel, as shown in formula (\ref{eq:kernel}).
\begin{equation}\label{eq:kernel}
    f_{\text{kernel}}=\frac{1}{N}\sum_{i}^{N} f_b[\text{pos}(i)]
\end{equation}
where $N$ denotes the number of user clicks, $\text{pos}(i)$ indexes the tokens positively clicked, and $[\cdot]$ denotes kernel selection in the dimension of the token length.

The kernel $f_{\text{kernel}}$ is regarded as the most representative feature capturing the user's intent.
In this case, tokens from the multi-scale representations similar to $f_{\text{kernel}}$ are considered important. Their importance is measured by the similarity with $f_{\text{kernel}}$, as shown in formula (\ref{eq:sim}).
\begin{equation}\label{eq:sim}
    s_j=\text{sigmoid}\left(\text{cos}(f_{\text{kernel}}, f_j)\right), j\in\{t, l\}
\end{equation}
where $\text{cos}(\cdot, \cdot)\in [-1, 1]$ gives the cosine similarity. Tokens with similar directions have a greater cosine distance, while those with dissimilar directions have a smaller cosine distance. The sigmoid function can be mapped the scores to range [0,1], which can be regarded as the importance of token in selection.
$j$ denotes the index of tokens of two different scales, including tiny (t) and large (l). 
$s_t\in\mathbb{R}^{L_t}$ and $s_l\in\mathbb{R}^{L_l}$ are similarity scores for the two scales.

After computing similarity scores between the multi-scale and base tokens, a differentiable top-k selection is proposed to individually select the the highest top $k_j$ score tokens, as shown in equation (\ref{eq:topk}).
\begin{equation}\label{eq:topk}
    s_{j}^{k_j}, \text{idx}_{j}^{k_j} = \text{torch.topk}(s_j)
\end{equation}
where torch.topk$(\cdot)$ is a Pytorch function, $s_j^{k_j}$ and $\text{idx}_{j}^{k_j}\in\mathbb{R}^{k_j}$ denote the top-k scores of different scale tokens.
Since these tokens vary in length, $k_j$ is generally set as 1/12 the length of the $L_j$.

Directly using $\text{idx}\_{j}^{k_j}$ for selection will impede gradient backpropagation. Therefore, we uses $s_j^{k_j}$ to keep differentiable.

Specifically, we convert $s_j^{k_j}$ into $k_j$ one-hot vectors of length $L_j$, and stack these vectors into a matrix $S_j^{k_j}=[I_{s_j^{1}}, I_{s_j^{2}}, \ldots,I_{s_j^{k_j}}]\in\mathbb{R}^{k_j\times L_j}$.

Afterward, the properties of matrix multiplication are used to choose the top $k_j$ tokens, as shown in equation (\ref{eq:selection}).
\begin{equation}\label{eq:selection}
    f_j^{k_j} = S_j^{k_j} f_{j}\in\mathbb{R}^{k_j\times C}
\end{equation}

To ensure the robustness of the network and its adaptability to multi-scale targets, the mean of the top-k scores $s_t^{k_t}$ and $s_l^{k_l}$ is calculated for two different scales. 
The scale with highest score will be selected to interact with $f_b$.

We use Cross Attention\cite{chen2021crossvit} to fuse multi-scale features, as illustrated in Fig. \ref{fig:multiscale}.
$f_b$ will be down-sampled using Scaled Cross-Attention, allowing for efficient updating of multi-scale features $f_j, j \in \{t, l\}$.

\subsection{Robust Token Selection based on Contrastive Loss}

In the proposed token selection algorithm, the similarity between the base and multi-scale tokens is crucial for selecting important tokens.

Ideally, the selected important tokens should belong to the target. 
However, many background tokens will also be wrongly selected, which can potentially contaminate the base tokens, as shown in the first row of Fig. \ref{fig:token_selection}.

To address this issue, a multi-scale token learning strategy based on contrastive learning is proposed.

During training, the positive and negative tokens in $f_t^{k_t}$ and $f_l^{k_l}$ will separated according to the formula (\ref{eq:contrast1}).
\begin{equation}\label{eq:contrast1}
\begin{split}
    f_j^{pos}&=f_j^{k_j}\odot y_j\\
    f_j^{neg}&=f_j^{k_j}\odot (1-y_j)
\end{split}
\end{equation}
where $j\in[t, l]$, $y_j$ is the ground truth to different scale tokens. The number of elements in $y_j$ equals the number of tokens in $f_j$.

After obtaining the positive and negative exemplar sets of important tokens, the goal is to maximize the similarity between $f_j^{pos}$ and $f_{\text{kernel}}$, while minimizing the similarity between $f_j^{neg}$ and $f_{\text{kernel}}$. This ensures that the important tokens are predominantly derived from target object.

The positively clicked tokens are denoted as $q=f_{\text{kernel}}$. The distance between $q$ and positive important tokens $k_j^+=f_j^{pos}$ is represented as $d_{pos}=|q-k_j^+|_2$, and the distance between $q$ and negative important tokens $k_j^-=f_j^{neg}$ is $d_{neg}=|q-k_j^-|_2$.

The objective is for distance $d_p=|q-k_j^+|$ between $q$ and $k_j^+$ to be lesser than the distance $d_n=|q-k_j^-|$ between $q$ and $k_j^-$. 
It can be achieved by using the loss function shown in equation (\ref{eq:contrast2}).

\begin{figure}[!t]
    \centering
    \includegraphics[scale=0.56]{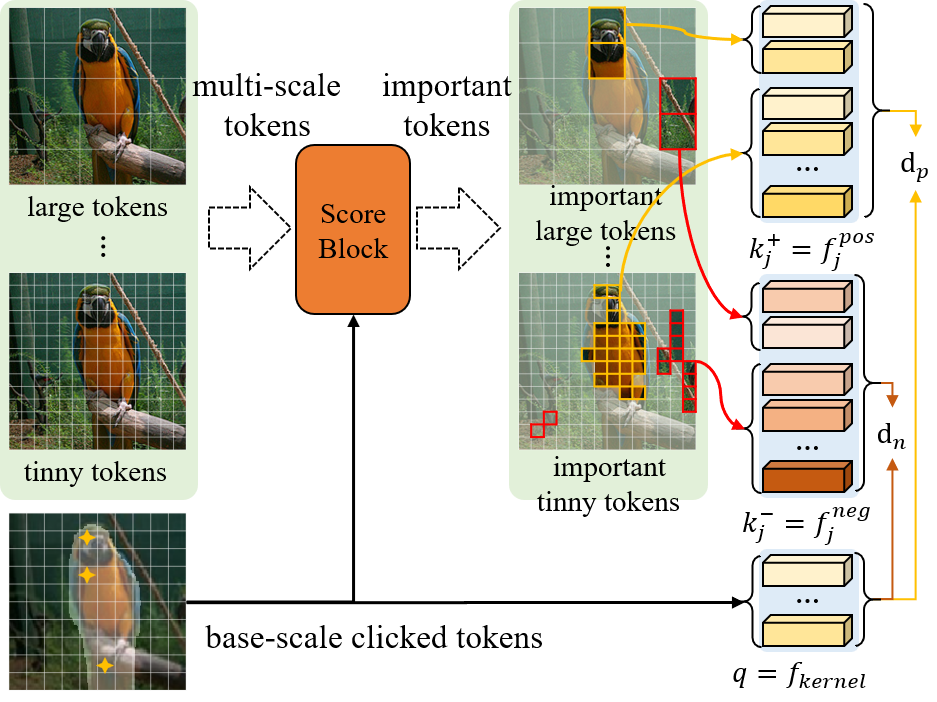}
    \caption{Triplet patch loss computation diagram. Red squares represent tokens that do not belong to the target, yellow squares represent tokens that belong to the target, and yellow asterisks represent user-clicked points}
    \label{fig:contrastloss}
\end{figure}

Another goal is maximizing the margin between $d_{neg}$ and $d_{pos}$.
This aids in imparting greater discriminative ability to the important tokens. 
Such discriminative ability can be obtained through the triplet patch loss function in formula (\ref{eq:contrast2}).
\begin{equation}\label{eq:contrast2}
\mathcal{L}_{c}=\sum_{j\in[t,l]}\frac{\log(1+\exp(d_{p}-d_{n}))}{N}
\end{equation}
where $N$ signifies the number of elements involved in the computation.
The loss function imposes a penalty when $d_p> d_n$, while the penalty $\mathcal{L}_{c}\rightarrow 0$ when $d_p << d_n$.
The overall workflow of the proposed loss function is illustrated in Fig. \ref{fig:contrastloss}.

In the model training phase, this loss is combined with the image segmentation loss for training.
\begin{equation}\label{eq:contrast3}
\mathcal{L}=\mathcal{L}_{seg}+\mathcal{L}_{c}
\end{equation}
where $\mathcal{L}_{seg}$ is the FocalLoss \cite{focalloss}.

\section{Experiment}
In this chapter, we first introduce the basic settings of the proposed algorithm as well as the training and validation strategy based on interactivity. 
Then, based on existing evaluation data, we compare the proposed algorithm with existing state-of-the-art (SOTA) algorithms in terms of accuracy and speed.
Afterward, we conduct ablation studies to validate the effectiveness of the proposed algorithm.
For ease of description, we denote the proposed algorithm as \textbf{MST} in the following sections.

\subsection{Experimental Configuration}

\noindent\textbf{Model selection.}
To satisfy the requirements for complex scenarios, SimpleClick \cite{liu2022simpleclick} is selected as the baseline framework of the proposed algorithm. 
Meanwhile, ViT-B \cite{dosovitskiy2020image} is adopted as the base image encoder for feature extraction.

\noindent\textbf{Training settings.}
In terms of training data generation, the random cropping image enhancement algorithm is adopted to generate training samples.
The size of training samples is $448\times 448$ and the training process is in the form of end-to-end.

Considered in light of the simulated strategy for click points, we employ the iterative learning strategy of RITM \cite{9897365}, at the same time, sampling positive and negative clicks through using the training data generation strategy proposed by Xu et al. \cite{7780416}.
To be specific, the maximum number of click points during training is set to 24, with a decay probability of 0.8.

For the purpose of getting better results, a combination of COCO \cite{lin2014microsoft} and LVIS \cite{gupta2019lvis} dataset is constructed for training the proposed algorithm and other comparable algorithmic approaches.
Additionally, the random flip and resize for data augmentation algorithms are adopted in training for generalization improvement.
The optimizer is adopted AdamW, $\beta_1=0.9,\beta_1=0.999$.

There are 30000 training samples in one epoch, with a total of 230 Epochs.
The initial learning rate is $5\times 10^{-6}$, and it is reduced by a factor of 10 when the epoch is 50 and 70.
The proposed algorithm is trained on six Nvidia RTX 3090 GPUs, which takes about 72 hours to complete.

\noindent\textbf{Evaluation strategy.}
With an aim of ensuring a consistent and fair evaluation, we choose to follow the techniques used for evaluation adopted by \cite{chen2021conditional, 8953578, 9157109, 9156403, 9897365, 7780416} to ensure a fair comparison.
During the process of evaluation, every point is sampled from the previous prediction locations which have the largest error, so as to ensure that the predicted IOU (Intersection Over Union) is as close to the target as possible.
The evaluation will be terminated when reached desired IOU or max click number.
The size of the image is $448\times 448$, which is the same as training samples.

For the evaluation metric, we employ the standard Number of Clicks (NoC) measure, revealing the average number of clicks needed to reach the predefined IOU (Intersection over Union).
Following previous works, the upper limit of the click number is 20. 
We report the model as a failure if it needs to reach the target IOU exceeding 20 clicks. 
Moreover, the NoF (Number of Failures) is also adopted in the proposed algorithm to evaluate the numbers of algorithm failures.

\subsection{Comparison with State-of-the-Art}
This section is concerned with making a comparison of the proposed algorithm and the current SOTA in terms of accuracy and complexity, with the effectiveness of the MST being further  analyzed.

\noindent\textbf{1) Overall performance on mainstream benchmarks. }

For obtaining a fair and better comparison, we do not compare the early classical methods such as Graph Cut \cite{boykov2001interactive} and Geodesic star convexity \cite{gulshan2010geodesic} with the proposed algorithm.
In this paper, SimpleClick-ViT-B \cite{liu2022simpleclick} is used as our baseline algorithm.
Currently, the high annotation quality of the LVIS dataset \cite{gupta2019lvis} has and the diversity of the COCO \cite{lin2014microsoft} dataset significantly make them become good choices for training the models. 

Hence, in this paper, all SOTA algorithms participating in the comparison are trained on the combined COCO-LVIS dataset.
The comparison results on mainstream benchmarks are illustrated in Table \ref{tab:sotacompar}.

\begin{table*}[!t]
\centering
\caption{Evaluation results on GrabCut, Berkeley, SBD, DAVIS, COCO Mval, and Pascal VOC. 'NoC 85/90' denotes the average Number of Clicks required the get IoU of 85/90\%. All methods are trained on COCO\cite{lin2014microsoft} and LVIS\cite{gupta2019lvis} datasets.\textcolor{red}{Red arrow}: MST $>$ 2n place. \textcolor{blue}{Blue arrow}: MST $>$ 3rd place. \textbf{Bold font}: best performance}
\label{tab:sotacompar}
\scalebox{0.78}{
\begin{tabular}{cccccccccc}
\hline
 & GrabCut \cite{rother2004grabcut} & Berkeley \cite{mcguinness2010comparative} & \multicolumn{2}{c}{SBD \cite{MCGUINNESS2010434}} & DAVIS \cite{perazzi2016benchmark} & \multicolumn{2}{c}{COCO\_MVal \cite{10.1007/978-3-319-10602-1_48}} & \multicolumn{2}{c}{PascalVOC \cite{everingham2009pascal}} \\
 & NoC 90 & NoC 90 & NoC 85 & NoC 90 & NoC 90 & NoC 85 & NoC 90 & NoC 85 & NoC 90 \\ \hline
f-BRS-B-hrnet32\cite{9156403} & 1.69 & 2.44 & 4.37 & 7.26 & 6.50 & - & - & -& - \\
RITM-hrnet18s\cite{9897365} & 1.68 & 2.60 & 4.25 & 6.84 & 5.98 & - & 3.58 & 2.57 & - \\
RITM-hrnet32\cite{9897365} & 1.56 & 2.10 & 3.59 & 5.71 & 5.34 & 2.18 & 3.03 & 2.21 & 2.59 \\
EdgeFlow-hrnet18\cite{Hao_2021_ICCV} & 1.72 & 2.40 & - & - & 5.77 & - & - & - & - \\
FocalClick-segformer-B0-S2\cite{chen2022focalclick} & 1.90 & 2.92 & 5.14 & 7.80 & 6.47 & 3.23 & 4.37 & 3.55 & 4.24\\
FocalClick-segformer-B3-S2\cite{chen2022focalclick} & 1.68 & 1.71 & 3.73 & 5.92 & 5.59 & 2.45 & 3.33 & 2.53 & 2.97\\
SAM-ViT-H\cite{kirillov2023segment} & 1.62 & 2.25 & 5.98 & 9.63 & 6.21 & 3.46 & 5.60 & 2.20 & 2.68\\
HQ-SAM-ViT-H\cite{sam_hq} & 1.84 & 2.00 & 6.23 & 9.66 & 5.58 & 3.81 & 5.94 & 2.50 & 2.93\\
SimpleClick-ViT-B\cite{liu2022simpleclick} & 1.48 & 1.97 & 3.43 & 5.62 & 5.06 & 2.18 & 2.92 & 2.06 & 2.38\\
\hline
Ours-ViT-B+MST &  1.52 & 1.77 & 3.03 & 5.20 & 4.82 & 2.11 & 2.88 & 2.05 & 2.37 \\
Ours-ViT-B+MST+CL & 1.48 &$\textbf{1.50}_{\textcolor{red}{12.28\% \uparrow}}$ & $\textbf{3.03}_{\textcolor{red}{11.6\%\uparrow}}$ & $\textbf{5.11}_{\textcolor{red}{9\% \uparrow}}$ & $\textbf{4.55}_{\textcolor{red}{10.07\%\uparrow}}$ & $\textbf{2.08}_{\textcolor{red}{4.59\%\uparrow}}$ & $\textbf{2.85}_{\textcolor{red}{2.4\%\uparrow}}$ & $\textbf{1.69}_{\textcolor{red}{17.9\%\uparrow}}$ & $\textbf{1.90}_{\textcolor{red}{20.16\%\uparrow}}$ \\
\hline
\end{tabular}
}
\end{table*}

In Table \ref{tab:sotacompar}, the MST and CL represent the proposed multi-scale tokens module and the contrastive loss-based token selection algorithm respectively.
Notably, it can be seen that the proposed MST model can effectively improve the performance of the baseline on different benchmarks.
This result can be attributed to the reason that the MST is capable of fully utilizing the multi-scale information of input and thus assisting the algorithm in performing well in scale variation cases.
Furthermore, CL plays an important role in helping the MST-based algorithm  reaching higher performance. This is mainly because similar tokens in the background cannot be constrained when only MST is used, while CL can ensure that the algorithm uses the multi-scale tokens of the target itself to interact with the base tokens as much as possible.

Along with demonstrating competitive performance in comparison to current popular segment anything (SAM) algorithms (SAM and HQ-SAM) the MST also achieves SOTA performance on all compared algorithms.

Additionally, compared with the current state-of-the-art algorithm SimpleClick on PascalVOC, the number of clicks is reduced by 20.16\%, and the performance on DAVIS and SBD is also enhanced by more than 10\%. 
In addition, the MST also outperforms the current SOTA algorithms FocalClick \cite{chen2022focalclick} and RITM \cite{9897365}.
Tellingly, this result verifies the effectiveness of the proposed algorithm.

Compared with the current popular segment anything (SAM) algorithms (SAM and HQ-SAM), the proposed algorithm also shows great performance advantages.

\noindent\textbf{2) Complexity Analysis}

The goal of the proposed algorithm is to use multi-scale tokens for achieving better performance in multi-scale scenarios.
Instead of directly applying multi-scale tokens for interaction, implementing the proposed algorithm is more efficient and accurate.

In order to intuitively evaluate the comparison between the MST, and the benchmark algorithm and the current advanced methods in terms of complexity, we count the parameter amount, calculation amount and inference speed of the algorithm respectively. 
The experimental results are listed in Table \ref{tab:speed_tab}.

\begin{table}[!t]
\centering
\caption{Efficiency comparison with SOTA. The inference speed is tested on GPU Nvidia 3090. (The speed of SAM and HQ-SAM are tested with an average of four clicks, one for encoding, and another for inference.)}
\label{tab:speed_tab}
\scalebox{0.8}{
\begin{tabular}{cccc}
\hline
Model Type & Params (MB) & FLOPs (G) & Speed/ms \\ \hline
SAM-ViT-H-1024 & 635.63 & 2802.69 &  1003\\
HQ-SAM-ViT-H-1024 & 637.23 & 2830.34 & 1102\\
SimpleClick-ViT-B-448 & 84.89 & 96.46 &  183\\
SimpleClick-ViT-L-448 & 322.18 & 266.44 & 300 \\
SimpleClick-ViT-H-448 & 659.39 & 700.96 & 585 \\ \hline
Ours-ViT-B+MST & 166.77 & 206.82 & 225 \\
\hline
\end{tabular}
}
\end{table}

Making comparisons between the MST and the baseline algorithm, we find that the parameters of the proposed algorithm have increased by 1.9 times and the calculation amount has increased by 2.1 times.
However, we also note that the running speed has only increased by 1.2 times. 
This is because we apply FlashAttention 2 \cite{dao2023flashattention} to speed up the attention operation.
Although our algorithm has increased complexity compared to the baseline algorithm, it brings considerable performance improvements and approaching inference speed.

Furthermore, it is notable that the MST requires fewer parameters when compared with SimpleClick-ViT-L and SimpleClick-ViT-H.
Besides, the MST also has faster inference speed.

As illustrated in Table \ref{tab:sotacompar}, the MST only needs 1/5 of the parameters and 1/10 of the calculation with a comparison to SAM-based algorithms including SAM and HQ-SAM.
At the same time, it is worth noting that SAM only needs to encode a specific image once, with the image encoding being processed in the decoding head with subsequent click operations. 
With a large number of clicks on the same image, its average speed will be quick.
To make fair comparisons,  we also average the results of four clicks when testing the speed of this type of algorithm.

Overall, although the numbers of parameters and calculations of the MST have increased  in comparison to the baseline algorithm, the reasoning speed of the MST can still maintain a relatively fast state due to the adoption of new technologies.
The fact that the proposed algorithm demonstrates faster inference speed and superior performance against models with larger parameters is hard to be ignored.

\subsection{Ablation Study}

The results of the ablation study conducted to evaluate the overall performance are presented in Table {\ref{tab:sotacompar}}.
This section introduces a new NoC-Scale evaluating indicator to evaluate the effectiveness of MST and CL compared to baseline SimpleClick-ViT-B.

\noindent\textbf{a). Effectiveness of multi-scale tokens}

\begin{figure*}[!ht]
    \centering
    \includegraphics[scale=0.46]{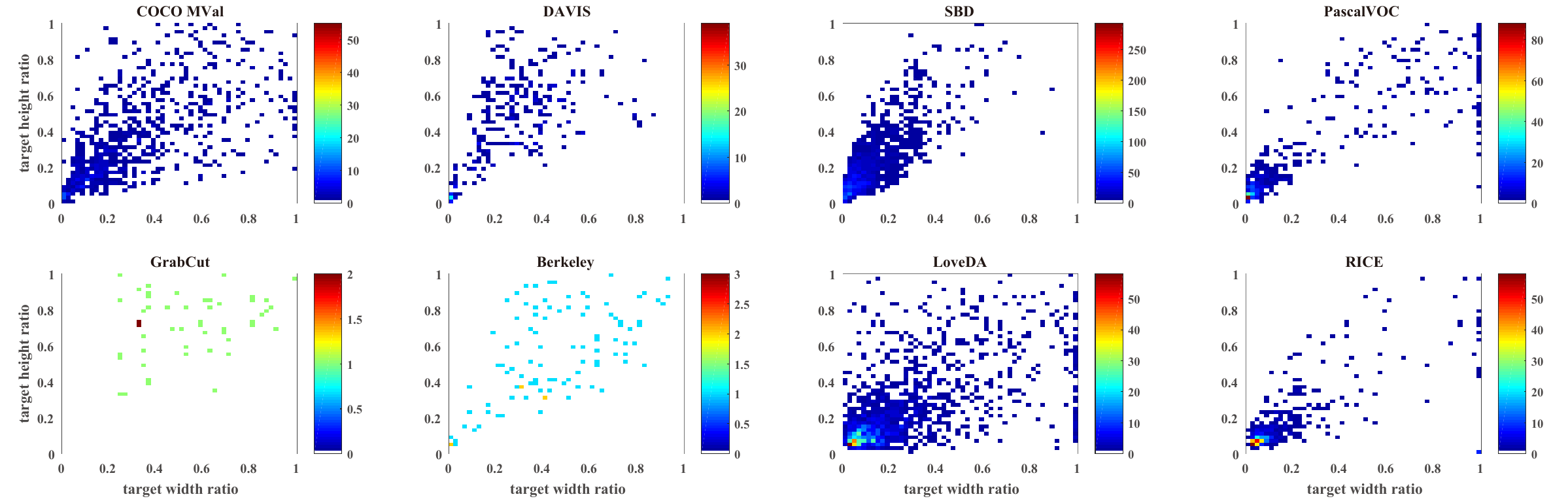}
    \caption{Distribution of object sizes in different interactive evaluation datasets. The horizontal axis represents the ratio of the target's width to the image's width, while the vertical axis represents the ratio of the target's height to the image's height.
    GrabCut, DAVIS, Berkeley, SBD, PascalVOC, and COCO\_MVal are image datasets of conventional scenes. LoveDA and RICE are datasets of remote-sensing images, both of which contain bird's-eye view images from satellites }
    \label{fig:scale_dist}
\end{figure*}

First, we conducted an analysis of the distribution of the target scale in the eight types of data sets, and the results are presented in Fig. \ref{fig:scale_dist}.

Fig. \ref{fig:scale_dist} displays the proportions of different target widths and heights in the dataset images, where a small proportion of width and height indicates a small target size, and a large proportion indicates a large target size.
It can be seen that the target size distribution in COCO\_MVal \cite{10.1007/978-3-319-10602-1_48} is relatively uniform, with the majority of targets occupying 0.2-0.3 of the total image size.
The size of the target in DAVIS \cite{perazzi2016benchmark} is basically concentrated between 0.3-0.8, suggesting a tendency for larger target sizes in this data set.
In contrast, the SBD \cite{MCGUINNESS2010434} dataset displays a concentration of target sizes between 0.1-0.3, indicating a prevalence of smaller target sizes.
In PascalVOC \cite{everingham2009pascal}, the distribution of target size is two extremes.Specifically, small targets (0.1-0.2) constitute the majority, and larger targets ($>$ 0.8) account for a significant portion, while the number of medium-sized targets (0.3-0.7) is relatively smaller.
It is noteworthy that the target size in GrabCut \cite{rother2004grabcut} and Berkeley \cite{mcguinness2010comparative} typically falls between 0.3-0.7, indicating a limited amount of data is small.

LoveDA \cite{loveda} and RICE \cite{lin2019remote} are remote sensing datasets.
The LoveDA dataset primarily consists of small-sized targets, and it also contains a large amount of large-sized targets because of the presence of various elements such as green land and water systems.

\begin{figure*}[!t]
    \centering
    \includegraphics[scale=0.65]{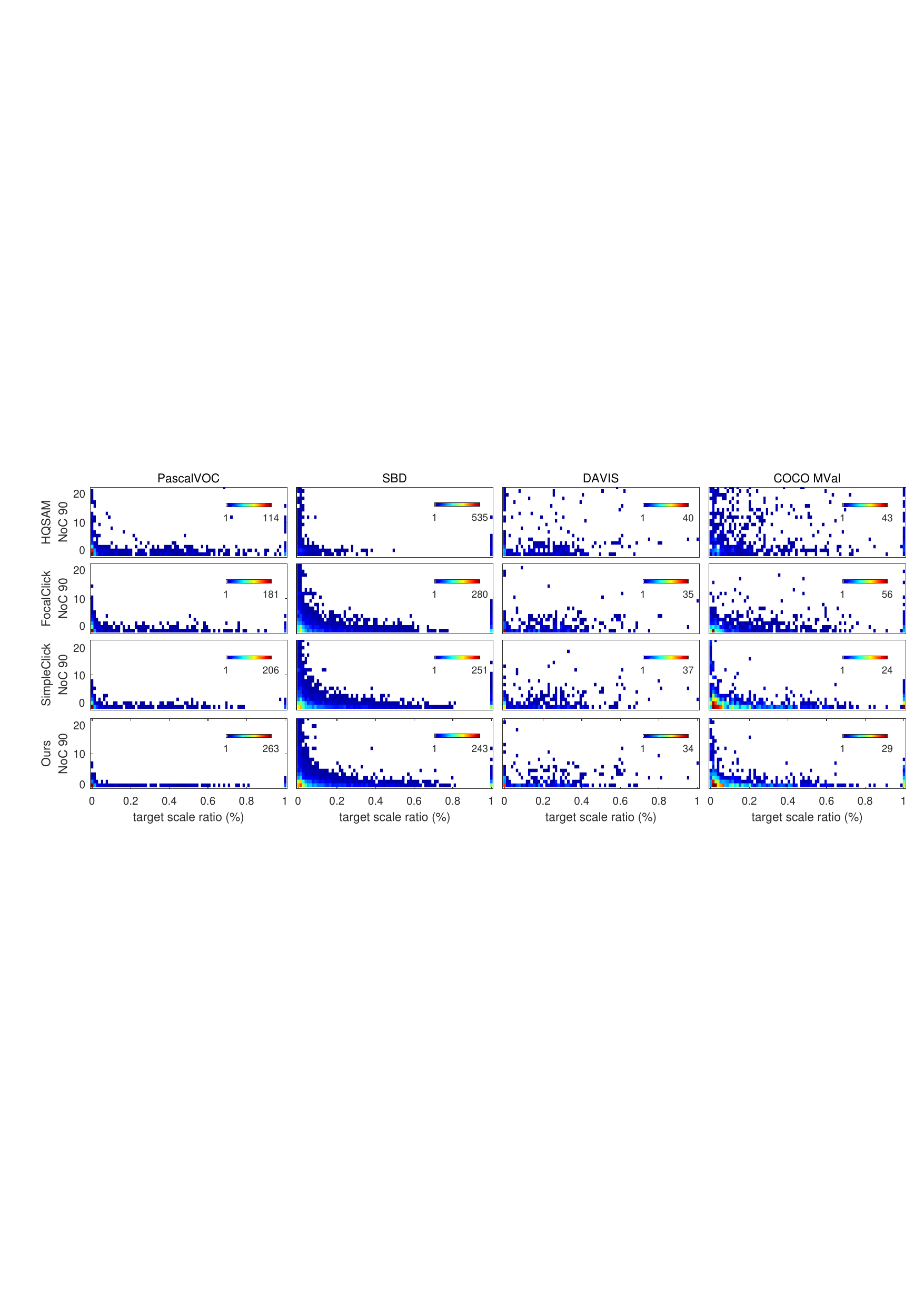}
    \caption{Distribution of NoC 90 of different algorithms on targets of different scales. The horizontal axis represents the ratio of the target area to the total image area, and the vertical axis represents the number of clicks (NoC) required to achieve target IOU 90\%. The different colors of the colorbar indicate the number of samples at the current location, with more samples corresponding to a redder color. It can be observed that a larger NoC generally occur when the target size is smaller, and there are also many instances where a larger NoC occurs in scenarios where the target area is very large. It can be seen that our algorithm performs excellently across various target sizes}
    \label{fig:nocvsscale}
\end{figure*}

To evaluate the effectiveness of the proposed MST in solving multi-scale changes, we analyzed the distribution of NoC 90 indicators at different scales of the algorithm, and the results are shown in Fig. \ref{fig:nocvsscale}.

It can be seen that the algorithm proposed in this paper can significantly reduce the NoC 90 on both large and small targets.

Therefore, the proposed MST effectively addresses the challenge of multi-scale variation in targets, which solidly verifies its effectiveness.

\noindent\textbf{b). Effectiveness of constructive loss in token selection}

\begin{figure*}[!ht]
    \centering
    \includegraphics[scale=0.58]{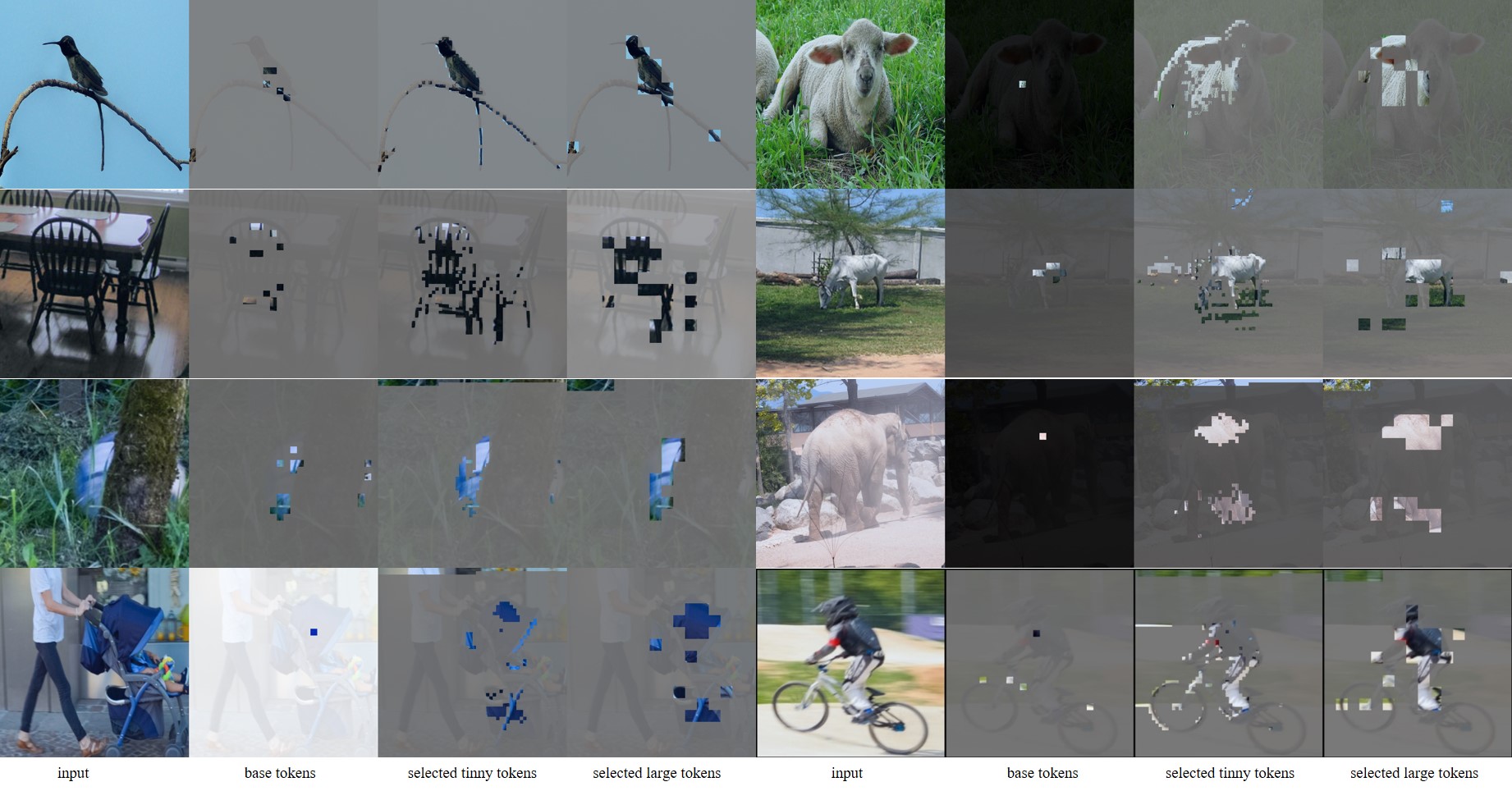}
    \caption{Results of multi-scale tokens selection. These tokens will be used to enhance the performance of base tokens in dealing with multi-scale variations. Selected tokens are represented by transparent squares, while unselected tokens are represented by opaque squares. The top two rows of image results show the effects of using contrastive loss, while the bottom two rows demonstrate the effects without the use of contrastive loss. It can be observed that after applying contrastive loss, most of the important tokens can be accurately selected. However, when contrastive loss is not used, there are some incorrectly selected tokens present in the edge regions of the images.}
    \label{fig:token_selection}
\end{figure*}

To evaluate the effectiveness of the contrastive loss in token selection, we evaluate the selection of tokens before and after using contrastive loss, and the experimental results are shown in Fig. \ref{fig:token_selection}.

In Fig. \ref{fig:token_selection}, the first and second rows represent the token selection results using contrastive loss, and the third and fourth lines represent the token selection results without using contrastive loss.

It can be seen that when contrastive loss is not used, although the algorithm can select tokens in most target regions, a large number of tokens in the background are selected at the same time. These target-independent tokens obviously affect the performance of the model, which is also verified by the overall ablation experimental results in Table \ref{tab:sotacompar}.

After using the contrastive loss, it can be seen that the accuracy of the selected tokens is significantly improved, and the selection of wrong tokens in the background is significantly suppressed.

This experimental result verifies the effectiveness of the proposed contrastive loss in improving the robustness of tokens.

\subsection{Performance for Mask Correction}
The mask correction task involves refining an initially provided mask with an IOU value between 0.75 and 0.85.
We evaluated the effectiveness of the proposed algorithm for mask correction using the DAVIS-585 dataset \cite{chen2022focalclick}. The experimental results are displayed in Table \ref{tab:maskcorrection}.

\begin{table}[!h]
\setlength{\tabcolsep}{3pt}
\centering
\caption{Quantitative results on DAVIS-585 benchmark. The metrics ‘NoC’ and ‘NoF’ mean the average Number of Clicks required and the Number of Failure examples for the target IOU.}
\label{tab:maskcorrection}
\scalebox{0.6}{
\begin{tabular}{cccc|ccc}
\hline
 & \multicolumn{3}{c|}{DAVIS585-SP} & \multicolumn{3}{c}{DAVIS585-ZERO} \\ 
 & NoC85 & NoC90 & NoF85 & NoC85 & NoC90 & NoF85 \\ \hline
RITM-hrnet18s\cite{9897365} & 3.71 & 5.96 & 49 & 5.34 & 7.57 & 52 \\
RITM-hrnet32\cite{9897365} & 3.68 & 5.57 & 46 & 4.74 & 6.74 & 45 \\
SimpleClick-ViT-B\cite{liu2022simpleclick} & 2.24 & 3.10 & 25 & 4.06 & 5.83 & 42 \\
SimpleClick-ViT-L\cite{liu2022simpleclick} & 1.81 & 2.57 & 25 & \textbf{3.39} & \textbf{4.88} & 36 \\
FocalClick-hrnet32-S2\cite{chen2022focalclick} & 2.32 & 3.09 & 28 & 4.77 & 6.84 & 48 \\
FocalClick-B3-S2\cite{chen2022focalclick} & 2.00 & 2.76 & 22 & 4.06 & 5.89 & 43 \\ \hline
Ours-ViT-B & \textbf{1.80}$_{\textcolor{red}{\uparrow{0.5\%}}}$ & \textbf{2.29}$_{\textcolor{red}{\uparrow{10.9\%}}}$ & \textbf{22}$_{\textcolor{red}{\uparrow{0\%}}}$ & 3.78$_{\textcolor{blue}{\uparrow{6.9\%}}}$ & 5.22$_{\textcolor{blue}{\uparrow{10.46\%}}}$ & 43 \\ \hline
\end{tabular}
}
\end{table}

It can be seen that, compared to RITM, SimpleClick, and FocalClick, MST has the best performance on the DAVIS585-SP task. 
Compared with the second-ranked SimpleClick-ViT-L, the proposed algorithm shows remarkable improvements of 10.9\% on NoC90, which proves the advanced nature of MST on mask correction tasks. 
Compared to SimpleClick-ViT-L, the proposed algorithm improves NoC 85 and NoC 90 in mask correction by 19.6\% and 26\% respectively, which demonstrates the effectiveness of the proposed algorithm in mask correction tasks.

The superior performance of our proposed algorithm can be attributed to its effective utilization of the input of the mask, which serves as a valuable source of certain prior information for the network. 
However, the existing algorithms only treat the mask as an additional modal feature, failing to fully make full use of it. 
In our study, we focus on calculating the similarity between the base tokens and other multi-scale tokens. 
In this way, the prior mask provides consistent feature expression for the target area tokens. So that it helps the algorithm to better find all similar tokens. Compared to other methods, MST can make better use of prior mask information.

On the DAVIS585-ZERO task, the proposed algorithm ranks second, only behind SimpleClick-ViT-L, whose parameter size is 1.94 times larger than ours. 
Compared to FocalClick, the proposed algorithm improves NoC 85 and NoC 90 by 6.9\% and 10.46\% respectively. 
This trend is consistent with the results shown in Table \ref{tab:sotacompar}.

\subsection{Qualitative Result}
\subsubsection{\textbf{One Click Performance}}
\begin{figure*}
    \centering
    \includegraphics[scale=0.6]{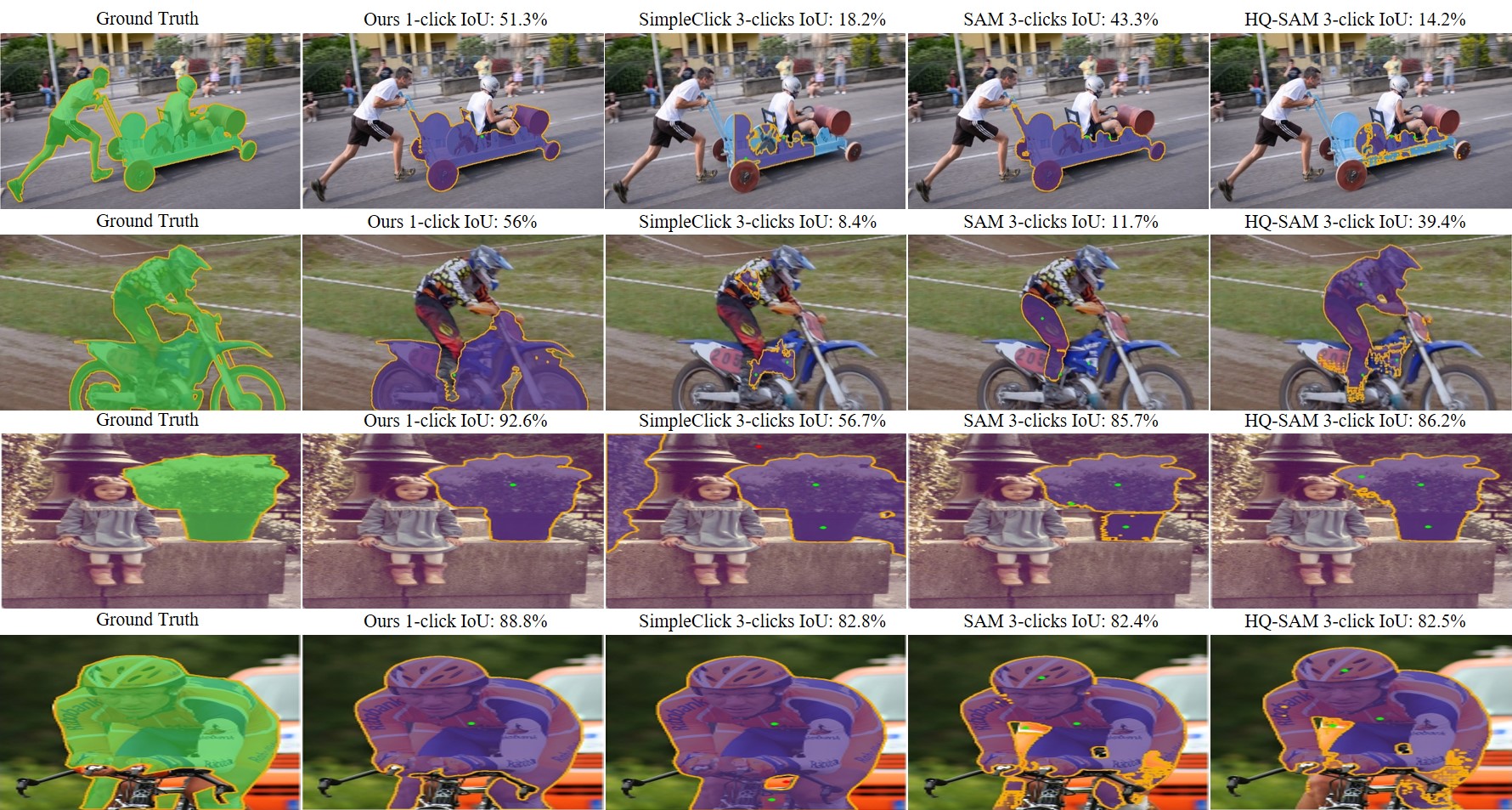}
    \caption{The proposed MST module increases the influence range of clicks, enabling it to achieve better performance with just a single click. The algorithms compared include the current state-of-the-art SimpleClick, SAM, and HQSAM. It is evident that in many instances, the performance of a single click from the proposed algorithm surpasses the performance of multiple clicks from the other algorithms}
    \label{fig:1click}
\end{figure*}

As shown in Fig. \ref{fig:miouclick}, the proposed algorithm has a great advantage in one-click performance.
The one-click experimental results are shown in Fig. \ref{fig:1click}.

In many cases, the proposed algorithm can achieve the target IOU in one click, which is significantly fewer clicks than required by other algorithms.
One possible reason is that the proposed MST module can effectively increase the influence range of user clicks, obtaining a larger receptive field than others.

\subsubsection{\textbf{Multi-Clicks Performance}}
\begin{figure*}
    \centering
    \includegraphics[scale=0.6]{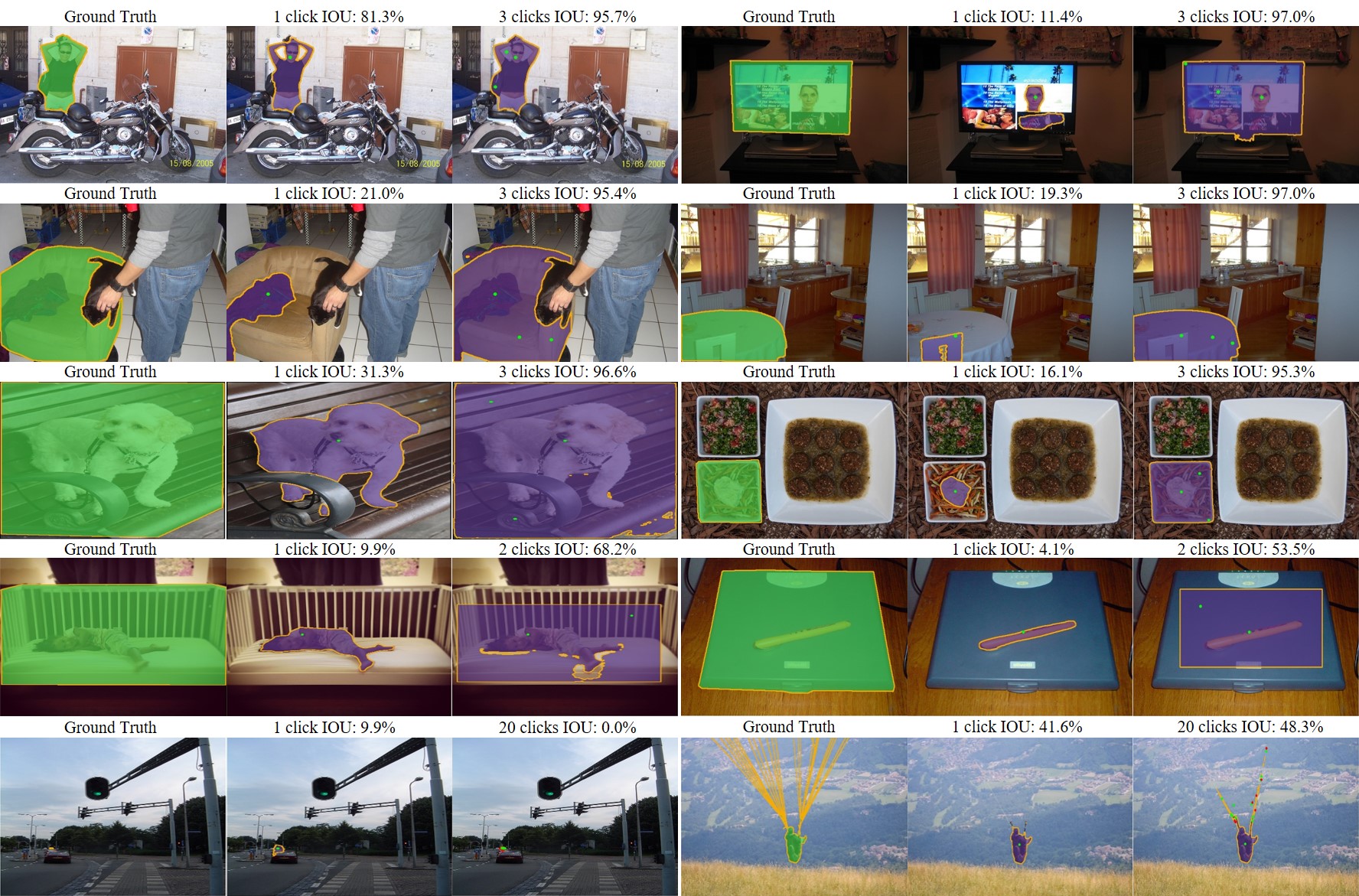}
    \caption{Quantitative analysis results of the proposed algorithm. The green mask denotes the ground truth, the purple mask denotes the prediction. These results are mainly used to analyze the effect of annotated data on algorithm performance. The first four rows are primarily used to analyze the differences between the algorithm's predicted results and the labels during the process of segmentation in electric motors. It can be seen that although the algorithm can segment a particular target well, this target may not correspond to the label, which could lead to inaccurate assessment results. The last row of images analyzes \textcolor{red}{the failure cases} of the algorithm proposed in this paper, showing that the performance of the proposed algorithm still requires further improvement when dealing with very small and intricate targets}
    \label{fig:quali_figs}
\end{figure*}

To evaluate multi-clicks performance, we analyzed the NoC results and \textbf{failure cases}, as shown in Fig. \ref{fig:quali_figs}.

As shown in Fig. \ref{fig:quali_figs}, the segmentation results of the proposed algorithm are finer than ground truth in some cases. 
For example, in the first image, the ground-truth label for the target is very rough, but the proposed algorithm can segment the target well. 
Some ground truth are mixed with other objects, which makes it hard for to obtain the 90\% IOU after 1 click.

In addition, the commonly used zoom-in operation also damages performance to a certain extent.
For example, after the second click on the 7th image, the predicted range is limited to a small rectangular area, which is much smaller than the ground truth.
In this case, more clicks are needed to enlarge the focus range of the algorithm to reach the target IOU.
The most intuitive evidence of this conclusion is the result in the 8th image. After the second click, due to the limited range of zoom-in, the IOU of the segmentation result cannot meet expectations, resulting in multiple clicks.

Furthermore, the proposed algorithm still needs to be improved in terms of the accuracy of object segmentation. For example, in the second image, although the algorithm can segment the portrait on the screen very well, the accuracy of the segmentation result is not fine enough, and even irrelevant objects are segmented.

\subsubsection{\textbf{Failure Cases}}
Finally, we show some cases that the proposed algorithm cannot handle, as shown in the last row of Fig. \ref{fig:quali_figs}.
These cases mainly occurred on very small and difficult-to-distinguish targets. In extreme cases, the IOU of the proposed algorithm is even 0.0\% after 20 times clicks.

\subsection{Generalization Evaluation on Remote Sensing Images}
In our study, all compared algorithms are trained on COCO-LVIS. However, COCO-LVIS' focus on common image types and scenes means it lacks generalization in some special visual data like remote sensing images.

In this section, we test all methods on the remote sensing dataset LoveDA \cite{loveda}, as shown in Table \ref{tab:rs_compar}.

\begin{table}[!h]
\caption{Comparison of algorithm generalization performance on LoveDA \cite{loveda}. None of the algorithms were trained on remote sensing images}
\label{tab:rs_compar}
\centering
\scalebox{0.8}{
\begin{tabular}{ccccc}
\hline
 & NoC 80 & NoC 85 & NoC 90 & NoF 90 \\ \hline
SAM-ViT-H & 11.96 & 14.61 & 17.42 & 1386 \\
HQ-SAM-ViT-H & 11.09 & 14.16 & 17.00 & 1346 \\
FocalClick-B3-S2 & 5.79 & 7.72 & 10.64 & 384 \\
SimpleClick-ViT-B & 5.91 & 7.65 & 10.44 & 302 \\
SimpleClick-ViT-L & 5.62 & 7.27 & 9.78 & 263 \\\hline
Ours & $\textbf{5.30}_{\textcolor{red}{5.69\%\uparrow}}$ & $\textbf{6.93}_{\textcolor{red}{4.7\%\uparrow}}$ & $\textbf{9.60}_{\textcolor{red}{1.8\%\uparrow}}$ & $\textbf{261}_{\textcolor{red}{0.76\%\uparrow}}$\\
\hline
\end{tabular}
}
\end{table}

Experimental results indicate that the proposed algorithm exhibits the best generalization performance on remote sensing data.

\section{Conclusion}
In this paper, we introduced a multi-scale token interaction algorithm (MST) to address the issue of target scale variation in interactive segmentation.
The MST algorithm used the user's clicks to derive similarity relationships among tokens of different scales, thereby facilitating extension to other vision transformer-based algorithms.
We also proposed a token selection strategy based on contrastive loss to increase the robustness of similarity computations, which significantly improved the quality of token selection.
Extensive experiments showed the effectiveness of the proposed algorithm.
Furthermore, we proposed a novel metric for evaluating multi-scale performance, which provided evidence of the algorithm's robustness and effectiveness.

\section{Acknowledgment}
This work was partially supported by Shenzhen Science and Technology Program under Grant JCYJ20210324115604012, Grant JCYJ20220818103006012, and Grant ZDSYS20220606100601002); in part by the Guangdong Basic and Applied Basic Research Foundation under Grant 2021B1515120008 and Grant 2023A1515011347; in part by Maxvision-AIRS-CUHK(SZ) Joint Laboratory of Inspection Robots; in part by Shenzhen Institute of Artificial Intelligence and Robotics for Society. Corresponding author: Yongquan Chen, (e-mail: yqchen@cuhk.edu.cn)

\bibliographystyle{IEEEtran}
\bibliography{refs.bib}

\end{document}